\documentclass[12pt]{article}
\usepackage[utf8]{inputenc}
\usepackage[margin=1in]{geometry}
\usepackage{amsmath,natbib}
\usepackage{amsfonts,mathrsfs,moreverb,lipsum}
\usepackage{graphicx,colonequals}
\usepackage{fixltx2e}
\usepackage{color}
\usepackage{caption}
\usepackage{subcaption}
\usepackage{pbox}
\newcommand{\vecX}{{\bf X}}
\newcommand{\bfx}{{\bf x}}
\newcommand{\bfw}{{\bf w}}

\newcommand{\bfz}{{\bf z}}
\newcommand{\bfZ}{{\bf Z}}
\newcommand{\bfwmu}{{\bf w}^{\mu}}
\newcommand{\bfwsig}{{\bf w}^{\Sigma}}
\newcommand{\wmu}{w^{\mu}}
\newcommand{\wsig}{w^{\Sigma}}
\newcommand{\rhomu}{\rho^{\mu}}
\newcommand{\brhomu}{{\boldsymbol \rho}^{\mu}}
\newcommand{\rhosig}{\rho^{\Sigma}}
\newcommand{\brhosig}{{\boldsymbol \rho}^{\Sigma}}
\newcommand{\Lmu}{L^{\mu}}
\newcommand{\Lsig}{L^{\Sigma}}
\newcommand{\lmu}{l^{\mu}}
\newcommand{\lsig}{l^{\Sigma}}
\newcommand{\btheta}{{\boldsymbol \Theta}}
\newcommand{\bvtheta}{{\boldsymbol \vartheta}}
\newcommand{\matsig}{{\boldsymbol \Sigma}}
\newcommand{\bLambda}{{\boldsymbol{\Lambda}}}

\newcommand{\matPsi}{\mathbf\Psi}
\newcommand{\ident}{\mathbf{I}}
\newcommand{\load}{\mathbf\Lambda}
\newcommand{\bpi}{{\boldsymbol \pi}}
\newcommand{\vecmu}{{\boldsymbol \mu}}
\title{Parameter-Wise Co-Clustering for High-Dimensional Data}
\author{M.P.B. Gallaugher$^*$, C. Biernacki$^{**}$, P.D. McNicholas$^*$}
\date{\small $^*$Department of Mathematics \& Statistics, McMaster University, Ontario, Canada.\\
$^{**}$Laboratory of Mathematics, University Lille 1, Lille, France.}

\begin{document}
\maketitle{}
\begin{abstract}
In recent years, data dimensionality has increasingly become a concern, leading to many parameter and dimension reduction techniques being proposed in the literature. A parameter-wise co-clustering model, for data modelled via continuous random variables, is presented. The proposed model, although allowing more flexibility, still maintains the very high degree of parsimony achieved by traditional co-clustering. A stochastic expectation-maximization (SEM) algorithm along with a Gibbs sampler is used for parameter estimation and an integrated complete log-likelihood criterion is used for model selection. Simulated and real datasets are used for illustration and comparison with traditional co-clustering.
\end{abstract}

\section{Introduction}
Clustering is the process of finding and analyzing underlying group structure in heterogenous data. With the emergence of big data, the number of variables in a dataset is constantly increasing and in many areas of application it is not uncommon for the number of variables to exceed the number of observations. In such situations, where the dimension of the data is very high, traditional mixture modelling techniques for clustering oftentimes fail.
Co-clustering is a very useful method for dealing with such scenarios. 

Co-clustering aims to define a partition in the rows of the data matrix for clustering individuals, as well as a partition in the columns for clustering variables. The result is partitioning the data matrix into homogenous blocks, or co-clusters, based on both individuals and variables. A key assumption for maintaining parsimony is that observations within each block are realizations of independent and identically distributed random variables. Some of the earliest work in co-clustering can be traced to \cite{hartigan72}. Since that time, model-based approaches have recently been shown to be effective for data treated as realizations of a continuous random variable \citep{nadif10}, count data \citep{pledger14} and ordinal data \citep{jacques17}, to name but a few.
In traditional co-clustering, added flexibility is often obtained by fitting more row-clusters and column-clusters; however, this is not generally advisable for parsimony reasons. Herein, we propose a parameter-wise co-clustering model that separately clusters columns according to both means and variances using the Gaussian distribution.

The remainder of this paper is laid out as follows. Section~2 presents a detailed background on high dimensional clustering techniques as well as details on traditional co-clustering using the Gaussian distribution. Section~3 presents the new model, parameter estimation, model selection criterion, and a non-exhaustive search algorithm for model selection. In Sections~4 and~5, synthetic and real datasets are considered for algorithm evaluation, classification performance, model selection performance, and comparison with traditional co-clustering. We conclude with a discussion of the results (Section~6).
\section{Gaussian-Based Clustering for High Dimensional Data}
\subsection{Model-Based Clustering}
Consider a dataset $\bfx=(\bfx_1',\bfx_2',\ldots,\bfx_n')'$ with $n$ individuals $\bfx_i\in \mathbb{R}^p$. One common method for clustering is model-based clustering, and generally makes use of a finite mixture model. A finite mixture model assumes that a real random vector $\vecX_i$ of dimension $p$ has probability density function 
$$
f(\bfx_i|{\boldsymbol \bvtheta})=\sum_{g=1}^G\pi_g f(\bfx_i|\btheta_g),
$$
where $\pi_g>0\ \forall\ g$ and $\sum_{g=1}^G\pi_g=1$ are the mixing proportions, $f(\cdot|\btheta_g)$ are the component density functions parameterized by $\btheta_g$, and $\bvtheta=(\pi_1,\ldots,\pi_G,\btheta_1,\ldots,\btheta_G)$ represents all the mixture parameters. 

Because of its mathematical tractability, the multivariate Gaussian mixture model is widely studied in the literature. In this case, each of the component densities is a multivariate Gaussian with density
$$
f(\bfx_i|\btheta_g)=\frac{1}{(2\pi)^{\frac{p}{2}}|\matsig_g|^{\frac{1}{2}}}\exp\left\{-\frac{1}{2}(\bfx_i-{\boldsymbol \mu}_g)'\matsig_g^{-1}(\bfx_i-{\boldsymbol \mu}_g)\right\},
$$ 
where $\btheta_g=({\boldsymbol \mu}_g,\matsig_g)$. The number of free parameters in a Gaussian mixture model is
\begin{equation}
\text{\#Params}_{\text{GaussMix}}=(G-1)+Gp+Gp(p+1)/2.
\label{eq:npgaussmix}
\end{equation}
Clearly, the number of free parameters in \eqref{eq:npgaussmix} is quadratic in the dimension of the data. As a result, using this simple mixture of Gaussian distributions will usually fail when the dimension $p$ increases.

In traditional model-based clustering, the group membership for observation $\bfx_i$ is usually represented by the vector $\bfz_i=(z_{i1},z_{i2},\ldots,z_{iG})$, where $z_{ig}=1$ if observation $\bfx_i$ belongs to group $g$ and 0 otherwise. Moreover, $\bfz_i$ is a realization of $\bfZ_i\sim\text{multinomial}(1;\bpi)$ where $\bpi=(\pi_1,\pi_2,\ldots,\pi_G)$. In addition, all couples $(\vecX_i,\bfZ_i)$ are usually assumed to be independent.
 
The use of a Gaussian mixture model for clustering can be traced back to \cite{wolfe65}. Other early work on Gaussian mixture models can be found in \cite{baum70} and \cite{scott71}. A detailed review of model-based clustering and classification is given by \cite{mcnicholas16b}, including related estimation and model selection procedures.

\subsection{High Dimensional Clustering Techniques}
Although the Gaussian mixture model is widely used, problems arise when the data dimensionality $p$ increases. The main contribution to the number of free parameters is through the component covariance matrices $\matsig_g$. Therefore, as a starting point, many methods try to impose parsimonious constraints on $\matsig_g$. A detailed background is presented by \cite{bouveyron14} and \cite{mcnicholas16b}. 

One particular example to note is the mixture of factor analyzers model. This model, presented by \cite{ghahramani97}, is a Gaussian mixture model with covariance structure
$\matsig_g=\bLambda_g\bLambda_g'+\matPsi$, where $\bLambda_g$ is a $p\times q$ matrix of factor loadings with $q<p$ and $\matPsi=\text{diag}(\psi_1,\psi_2,\ldots,\psi_p)$, $\psi_j\in\mathbb{R}^+$. Numerous extensions are proposed in the literature, including \cite{mclachlan00a}, who utilize the more general structure $\matsig_g=\bLambda_g\bLambda_g'+\matPsi_g$, and the closely-related mixture of probabilistic principal component analyzers with $\matsig_g=\bLambda_g\bLambda_g'+\psi_g\ident$ \citep{tipping99b}. In addition to these minor extensions, \cite{mcnicholas08} construct a family of eight parsimonious Gaussian models by considering the constraint $\load_g=\load$ in addition to $\matPsi_g=\matPsi$ and $\matPsi_g=\psi_g\ident$.  For the fully constrained model in \cite{mcnicholas08}, there are
\begin{equation}
\text{\#Params}_{\text{MFA}}=(G-1)+Gp+pq-q(q-1)/2+1
\end{equation}
free parameters. It is clear that although the number of free parameters associated with these models is linear in $p$, it is still nevertheless dependent on the dimension. Consequently, these models are still not suitable for very high dimensional data. Moreover, these methods may not be viable when $n>p$, which is common in applications such as gene expression data, word processing data, single nucleotide polymorphism data, etc.  

Alternatively, \cite{bouveyron07a} use the spectral decomposition of $\matsig_g$, i.e., 
$
\matsig_g={\bf D}_g{\boldsymbol \Delta}_g{\bf D}'_g,
$
where ${\bf D}_g$ is the orthogonal matrix of eigenvectors and ${\boldsymbol \Delta}_g$ is a diagonal matrix of corresponding eigenvalues for which they impose the structure $${\boldsymbol \Delta}_g=\text{diag}(a_{1g},a_{2g},\ldots,a_{q_gg},b_g,b_g,\ldots,b_g),$$ where $a_{kg}$ are the $q_g$ largest eigenvalues and $b_g$ is average of the remaining $p-q_g$ eigenvalues. This also greatly reduces the number of free parameters, i.e.,
\begin{equation}
\text{\#Params}_{\text {Bouveyron}}=(G-1)+Gp+\sum_{g=1}^G q_g[p-(q_g+1)/2]+\sum_{g=1}^G q_g+2G.
\end{equation}
Again, however, the number of free parameters is dependent on the dimensionality of the data.

Finally, there are also variable selection procedures such as $\ell_1$ penalization methods which take advantage of sparsity to perform variable selection and parameter estimation simultaneously. The first such proposed method is presented by \cite{pan07} who consider equal, diagonal covariance matrices between groups and apply an $\ell_1$ penalty to the mean vectors. A lasso method is then used for parameter estimation. This is extended by \cite{zhou09}, who consider unconstrained covariance matrices and apply an $\ell_1$ penalty for both the mean and covariance parameters. Although these methods are useful for dealing with the dimensionality problem, the $\ell_1$ penalty shrinks the parameters, thus introducing bias, as discussed by \cite{meynet12}. Moreover, the Bayesian information criterion \citep[BIC;][]{schwarz78} may not be suitable for high-dimensional data. A detailed review of each of these methods is given by \cite{biernacki17}.
  
\subsection{Co-Clustering and its Limitations}
Co-Clustering is a very useful tool for analyzing high-dimensional data. This method considers simultaneous partitions of rows and columns, which are then used to organize the data into homogenous blocks. For traditional co-clustering, as in clustering, data are assumed to come in the form of an $n\times p$ matrix $\bfx$ with rows represented by $\bfx_i'$. Each individual element of $\bfx_i$ is denoted by $x_{ij}$, so that $x_{ij}$ is the observation in row $i$ and column $j$. 

In co-clustering, there is an unknown partition of the rows into $G$ clusters, from this point onwards referred to as row-clusters, represented by the indicator vector $\bfz_i$ as defined previously. Unlike traditional co-clustering, however, there is also a partition of the columns into $L$ clusters, referred to as column-clusters, represented by the indicator vector $\bfw_j=(w_{j1},w_{j2},\ldots,w_{jL})\sim\text{multinomial}(1;{\boldsymbol \rho})$, where $w_{jl}=1$ if column $j$ belongs to column-cluster $l$ and $w_{jl}=0$ otherwise, and ${\boldsymbol \rho}=(\rho_1,\rho_2,\ldots,\rho_L)$. It is assumed that each data point $x_{ij}$ is independent once the  $\bfz_i$ and $\bfw_j$ are fixed. If, in addition, all $\bfz_i$ and $\bfw_j$ are assumed independent, and the latent block model is utilized in the same manner as \cite{nadif10}, then the joint density of $\bfx$ becomes $
f(\bfx;\bvtheta)=\sum_{\bfz\in \mathcal{Z}}\sum_{\bfw\in \mathcal{W}}p(\bfz;{\boldsymbol \pi})p(\bfw;{\boldsymbol \rho})f(\bfx|\bfz,\bfw;\btheta),$
where 
\begin{equation*}\begin{split}
&p(\bfz;{\boldsymbol \pi})=\prod_{i=1}^n\prod_{g=1}^G\pi_g^{z_{ig}},\qquad 
p(\bfw;{\boldsymbol \rho})=\prod_{j=1}^p\prod_{l=1}^{L}{\rho_{l}}^{w_{jl}}, \qquad\text{and}\\
&f(\bfx|\bfz,\bfwmu,\bfwsig;\btheta)=\prod_{i=1}^n\prod_{g=1}^G\prod_{j=1}^d\prod_{l=1}^{L}\left[\frac{1}{\sqrt{2\pi}\sigma_{gl}}\exp\left\{-\frac{1}{2\sigma^2_{gl}}(x_{ij}-\mu_{gl})^2\right\}\right]^{z_{ig}w_{jl}},
\end{split}\end{equation*}
where $\mu_{gl}$ and $\sigma^2_{gl}$ are the mean and variance, respectively, for row-cluster $g$ and column-cluster $l$, $\btheta$ is the set of all $\mu_{gl}$ and $\sigma^2_{gl}$, and $\bvtheta=(\bpi,{\boldsymbol \rho},\btheta)$. The total number of free parameters in this traditional co-clustering model is \begin{equation}\label{eqn:ccparas}\text{\#Params}_{\text{trad coclust}}=G+L+2(GL-1).\end{equation} Note that \eqref{eqn:ccparas} does not depend on the dimension, making it a very parsimonious model. Moreover, co-clustering is still possible to perform when $p>n$.

There are two different ways that one can view co-clustering. The first is that the main goal is the clustering of rows, and the clustering of columns is solely a way to solve the problem of dimensionality. However, in certain applications, the clustering of the columns might also be of interest.

\paragraph{Limitations of Co-Clustering}Although co-clustering has advantages over other high dimensional techniques (especially in the number of free parameters), the model is fairly restrictive because all observations in a block are realizations of independent and identically distributed Gaussian random variables with mean $\mu_{gl}$ and variance $\sigma^2_{gl}$. More flexibility is obtained by fitting more column-clusters and row-clusters, which is not always possible or advisable. What we propose in the present work is a parameter-wise co-clustering method by clustering columns according to both means and variances. This is the reason why we adopt hereafter the denomination ``parameter-wise" co-clustering, which is now presented in detail.

\section{Parameter-Wise Gaussian Co-Clustering}
\subsection{A Model to Combine Two Latent Variables in Columns}
Recall that traditional co-clustering aims to cluster data such that observations in the same block have the same distribution. An extension of traditional co-clustering for data treated as realizations of a Gaussian random variable is now considered. Similar to traditional co-clustering, there is a partition in rows and columns. However, now there are two partitions in the columns; specifically, a partition with respect to means and a partition with respect to variances.

Recall also that the data, which are treated as realizations of a continuous random variable, are represented as an $n\times p$ matrix, $\bfx=(x_{ij})_{1\le i\le n,1\le j\le p}$. The partition in rows is again represented by $\bfz=(\bfz_1,\bfz_2,\ldots,\bfz_n)$. 

\paragraph{Two Partitions in Columns}
The partition in columns by means is represented by $\bfwmu=(\bfwmu_1,\bfwmu_2,\ldots,\bfwmu_p)$, where 
$$
\bfwmu_j=(\wmu_{j1},\wmu_{j2},\ldots,\wmu_{j\Lmu})\sim\text{multinomial}(1;\brhomu)
$$
with $\brhomu=(\rhomu_1,\rhomu_2,\ldots,\rhomu_{\Lmu})$ and the partition in columns by variances is denoted by $\bfwsig=(\bfwsig_1,\bfwsig_2,\ldots,\bfwsig_p)$, where
$$
\bfwsig_j=(\wsig_{j1},\wsig_{j2},\ldots,\wsig_{j\Lsig})\sim\text{multinomial}(1;\brhosig)
$$
with $\brhosig=(\rhosig_1,\rhosig_2,\ldots,\rhosig_{\Lsig})$.
These two partitions in the columns is where the main novelty lies.
Note that $G,\Lmu$ and $\Lsig$ are the number of row-clusters, column-clusters by means, and column-clusters by variances, respectively. 
\paragraph{Log-Likelihood}
Using a small extension of the latent block model the observed log-likelihood is then
$$
f(\bfx;\bvtheta)=\sum_{\bfz\in \mathcal{Z}}\sum_{\bfwmu\in \mathcal{W}^{\mu}}\sum_{\bfwsig\in \mathcal{W}^{\Sigma}}p(\bfz;{\boldsymbol \pi})p(\bfwmu;\brhomu)p(\bfwsig;\brhosig)f(\bfx|\bfz,\bfwmu,\bfwsig;\vecmu,\matsig),
$$
where
\begin{equation*}\begin{split}
&p(\bfz;{\boldsymbol \pi})=\prod_{i=1}^n\prod_{g=1}^G\pi_g^{z_{ig}},\quad
p(\bfwmu;\brhomu)=\prod_{j=1}^p\prod_{\lmu=1}^{\Lmu}{(\rhomu_{\lmu})}^{\wmu_{j\lmu}},\quad
p(\bfwsig;\brhosig)=\prod_{j=1}^p\prod_{\lsig=1}^{\Lsig}{(\rhosig_{\lsig})}^{\wsig_{j\lsig}},\quad\text{and}\\
&f(\bfx|\bfz,\bfwmu,\bfwsig;\vecmu,\matsig)=\prod_{i=1}^n\prod_{g=1}^G\prod_{j=1}^p\prod_{\lmu=1}^{\Lmu}\prod_{\lsig=1}^{\Lsig}\left[\frac{1}{\sqrt{2\pi}\sigma_{g\lsig}}\exp\left\{-\frac{1}{2\sigma^2_{g\lsig}}(x_{ij}-\mu_{g\lmu})^2\right\}\right]^{z_{ig}\wmu_{j\lmu}\wsig_{j\lsig}}.
\end{split}\end{equation*}
In terms of notation, $\vecmu=(\vecmu_1,\vecmu_2,\ldots,\vecmu_G)$, where $\vecmu_g=(\mu_{g1},\mu_{g2},\ldots,\mu_{g\Lmu})$. Note that $\mu_{g\lmu}$ is the mean for row-cluster $g$ and column-cluster by means $\lmu$. Likewise, $\matsig=(\matsig_1,\matsig_2,\ldots,\matsig_G)$, where $\matsig_{g}=(\sigma_{g1}^2,\sigma_{g2}^2,\ldots,\sigma_{g\Lsig}^2)$ and $\sigma_{g\lsig}^2$ is the variance for row-cluster $g$ and column-cluster by variances $\lsig$.
Finally, the complete-data log-likelihood is
\begin{equation*}\begin{split}
p(\bfx,\bfz,\bfwmu,\bfwsig;{\boldsymbol \bvtheta})=C+\sum_{i=1}^n&\sum_{g=1}^Gz_{ig}\log\pi_g+\sum_{j=1}^p\sum_{\lmu=1}^{\Lmu}\wmu_{j\lmu}\log\rhomu_{\lmu}+\sum_{j=1}^p\sum_{\lsig=1}^{\Lsig}\wsig_{j\lsig}\log\rhosig_{\lsig}\\
&-\frac{1}{2}\sum_{i=1}^n\sum_{g=1}^G\sum_{j=1}^p\sum_{\lmu=1}^{\Lmu}\sum_{\lsig=1}^{\Lsig}z_{ig}\wmu_{j\lmu}\wsig_{j\lsig}\left[\log\sigma^2_{g\lsig}+\frac{(x_{ij}-\mu_{g\lmu})^2}{\sigma^2_{g\lsig}}\right],
\end{split}\end{equation*}
where $C$ is a constant with respect to the parameters and $\bvtheta=(\bpi,\brhomu,\brhosig,\vecmu,\matsig)$. From this point on, we refer to this model as parameter-wise co-clustering.

\paragraph{Number of Free Parameters}
The number of free parameters in the parameter-wise co-clustering model is
\begin{equation*}\begin{split}
\text{\#Params}_{\text {new coclust}}&=G-1+\Lmu-1+\Lsig-1+G\Lmu+G\Lsig\\
&=G+(\Lmu+\Lsig)(G+1)-3.
\end{split}\end{equation*}
There are a few comparisons with traditional co-clustering that are now discussed. First, similar to traditional co-clustering, the number of free parameters for the proposed parameter-wise method is independent of the dimension, meaning a high degree of parsimony is still maintained. Before mentioning the second point, note that the column-clusters by means and column-clusters by variances can be combined. For example, columns in column-cluster~1 by means and column-cluster~1 by variances can be combined to form one column-cluster. In general, columns in column-cluster $\lmu$ by means and column-cluster $\lsig$ by variances can be combined to form one column-cluster for any combination of $\lmu$ and $\lsig$, leading to a maximum of $\Lmu\Lsig$ column-clusters. There can, however, be fewer than $\Lmu\Lsig$ combined column-clusters because it is possible, for example, that no columns are clustered into column-cluster~3 by means and column-cluster~2 by variances. Now, assuming $G$ is equal for both parameter-wise and traditional co-clustering, and $\Lmu=\Lsig=L$, then there are only an additional $L-1$ free parameters when using the parameter-wise model. Although there are these additional free parameters, there is the possibility of $L^2$ combined column-clusters, allowing for a finer partition of the columns and increased flexibility.

There is also the possibility that the parameter-wise model has fewer free parameters than traditional co-clustering while still maintaining similar flexibility. For example, if traditional co-clustering is considered with $G=4$ and $L=5$, then the total number of free parameters is 47. In the parameter-wise case, if $G=4$, $\Lmu=3$, $\Lsig=3$, then the total number of free parameters is 31. In this case, there is a possibility of a total of nine column-clusters compared to five column-clusters when using traditional co-clustering.

\subsection{Parameter Estimation Using the SEM Gibbs Algorithm}
The SEM algorithm after initialization at iteration $q$ proceeds as follows.\\[+2pt]
{\bf SE Step}:
Generate the row partition $\bfz^{(q+1)}$ according to 
$$
P(z_{ig}=1|\bfx,{\bfwmu}^{(q)},{\bfwsig}^{(q)};\vecmu^{(q)},\matsig^{(q)},\bpi^{(q)})=\frac{\pi_g^{(q)}f(\bfx_i|{\bfwmu}^{(q)},{\bfwsig}^{(q)};\vecmu_g^{(q)},\matsig_g^{(q)})}{\sum_{g'}^G\pi_{g'}^{(q)}f(\bfx_i|{\bfwmu}^{(q)},{\bfwsig}^{(q)};\vecmu_{g'}^{(q)},\matsig_{g'}^{(q)})},
$$
where 
$$
f(\bfx_i|{\bfwmu}^{(q)},{\bfwsig}^{(q)};\vecmu_g^{(q)},\matsig_g^{(q)})=\prod_{j=1}^p\prod_{\lmu=1}^{\Lmu}\prod_{\lsig=1}^{\Lsig}\left[\frac{1}{\sqrt{2\pi}\sigma^{(q)}_{g\lsig}}\exp\left\{-\frac{1}{2{\sigma^2}^{(q)}_{g\lsig}}(x_{ij}-\mu^{(q)}_{g\lmu})^2\right\}\right]^{{\wmu}^{(q)}_{j\lmu}{\wsig}^{(q)}_{j\lsig}}.
$$
Generate the column partition by means ${\bfwmu}^{(q+1)}$ according to
$$
P(\wmu_{j\lmu}=1|\bfx,{\bfz}^{(q+1)},{\bfwsig}^{(q)};\vecmu^{(q)},\matsig^{(q)},{\brhomu}^{(q)})=\frac{{\rhomu}^{(q)}_{\lmu} f(\bfx_{\cdot j}|{\bfz}^{(q+1)},{\bfwsig}^{(q)};\vecmu_{\lmu}^{(q)},\matsig^{(q)})}{\sum_{{\lmu}'}^{\Lmu}{\rhomu}^{(q)}_{{\lmu}'} f(\bfx_{\cdot j}|{\bfz}^{(q+1)},{\bfwsig}^{(q)};\vecmu_{{\lmu}'}^{(q)},\matsig^{(q)})},
$$
where $\bfx_{\cdot j}=(x_{1j},x_{2j},\ldots,x_{nj})$, $\vecmu_{\lmu}^{(q)}=(\mu_{1\lmu}^{(q)},\mu_{2\lmu}^{(q)},\ldots,\mu_{G\lmu}^{(q)})$, and
$$
f(\bfx_{\cdot j}|{\bfz}^{(q+1)},{\bfwsig}^{(q)};\vecmu_{\lmu}^{(q)},\matsig^{(q)})=\prod_{i=1}^n\prod_{g=1}^G\prod_{\lsig=1}^{\Lsig}\left[\frac{1}{\sqrt{2\pi}\sigma^{(q)}_{g\lsig}}\exp\left\{-\frac{1}{2{\sigma^2}^{(q)}_{g\lsig}}(x_{ij}-\mu^{(q)}_{g\lmu})^2\right\}\right]^{z_{ig}^{(q+1)}{\wsig}^{(q)}_{j\lsig}}.
$$
Generate the column partition by variances ${\bfwsig}^{(q+1)}$ according to
$$
P(\wsig_{j\lsig}=1|\bfx,{\bfz}^{(q+1)},{\bfwmu}^{(q+1)};\vecmu^{(q)},\matsig^{(q)},{\brhosig}^{(q)})=\frac{{\rhosig}^{(q)}_{\lsig} f(\bfx_{\cdot j}|{\bfz}^{(q+1)},{\bfwmu}^{(q+1)};\vecmu^{(q)},\matsig_{\lsig}^{(q)})}{\sum_{{\lsig}'}^{\Lsig}{\rhosig}^{(q)}_{{\lsig}'} f(\bfx_{\cdot j}|{\bfz}^{(q+1)},{\bfwmu}^{(q+1)};\vecmu^{(q)},\matsig_{{\lsig}'}^{(q)})},
$$
where $\matsig_{\lsig}^{(q)}=({\sigma^2}_{1\lsig}^{(q)},{\sigma^2}_{2\lsig}^{(q)},\ldots,{\sigma^2}_{G\lsig}^{(q)})$ and
$$
f(\bfx_{\cdot j}|{\bfz}^{(q+1)},{\bfwmu}^{(q+1)};\vecmu^{(q)},\matsig_{\lsig}^{(q)})=\prod_{i=1}^n\prod_{g=1}^G\prod_{\lmu=1}^{\Lmu}\left[\frac{1}{\sqrt{2\pi}\sigma^{(q)}_{g\lsig}}\exp\left\{-\frac{1}{2{\sigma^2}^{(q)}_{g\lsig}}(x_{ij}-\mu^{(q)}_{g\lmu})^2\right\}\right]^{z_{ig}^{(q+1)}{\wmu}^{(q+1)}_{j\lmu}}.
$$
{\bf M Step}: Update the parameters according to
$$
\pi_{g}^{(q+1)}=\frac{\sum_{i=1}^nz_{ig}^{(q+1)}}{n}, \qquad {\rhomu_{\lmu}}^{(q+1)}=\frac{\sum_{j=1}^p{\wmu_{j\lmu}}^{(q+1)}}{p}, \qquad {\rhosig_{\lsig}}^{(q+1)}=\frac{\sum_{j=1}^p{\wsig_{j\lsig}}^{(q+1)}}{p},
$$
\begin{equation*}\begin{split}
\mu_{g\lmu}^{(q+1)}&=\frac{\sum_{i=1}^n\sum_{j=1}^p\sum_{\lsig=1}^{\Lsig}z_{ig}^{(q+1)}{\wmu_{j\lmu}}^{(q+1)}{\wsig_{j\lsig}}^{(q+1)}x_{ij}}{\sum_{i=1}^n\sum_{j=1}^p\sum_{\lsig=1}^{\Lsig}z_{ig}^{(q+1)}{\wmu_{j\lmu}}^{(q+1)}{\wsig_{j\lsig}}^{(q+1)}}=\frac{\sum_{i=1}^n\sum_{j=1}^pz_{ig}^{(q+1)}{\wmu_{j\lmu}}^{(q+1)}x_{ij}}{\sum_{i=1}^n\sum_{j=1}^pz_{ig}^{(q+1)}{\wmu_{j\lmu}}^{(q+1)}},
\end{split}\end{equation*}

$$
{\sigma^2_{g\lsig}}^{(q+1)}=\frac{\sum_{i=1}^n\sum_{j=1}^p\sum_{\lmu=1}^{\Lmu}z_{ig}^{(q+1)}{\wmu_{j\lmu}}^{(q+1)}{\wsig_{j\lsig}}^{(q+1)}(x_{ij}-\mu_{g\lmu}^{(q+1)})^2}{\sum_{i=1}^n\sum_{j=1}^p\sum_{\lmu=1}^{\Lmu}z_{ig}^{(q+1)}{\wmu_{j\lmu}}^{(q+1)}{\wsig_{j\lsig}}^{(q+1)}}.
$$

After a burn-in period of the algorithm, the estimates of each of the parameters are just the mean of the runs of the SEM algorithm (the number of runs are assessed experimentally in Section~4). We denote these final estimates by $\hat{\bvtheta}=(\hat{\bpi},\hat{\brhomu},\hat{\brhosig},\hat{\vecmu},\hat{\matsig})$. For the final partition of rows, columns by means, and columns by variances, we fix the parameters at their estimates and run more iterations of the SE step. We then assign each row to the row-cluster to which it is assigned most often over these additional SE steps. Likewise, each column is assigned to the column-cluster by means to which it is assigned most often over the additional SE steps, and finally each column is assigned to the column-cluster by variances to which it is assigned most often over the additional SE iterations. For our simulations and real data analyses, we take 20 such runs to obtain the final partitions $\hat{\bfz},\hat{\bfwmu}$, and $\hat{\bfwsig}$.

\subsection{Model Selection}
\paragraph{ICL--BIC}
As is the case in any clustering scenario, the number of row-clusters, column-clusters by means, and column-clusters by variances are not known {\it a priori} and, therefore, a model selection criterion is required. Similar to traditional co-clustering, the observed log-likelihood is intractable and so the BIC cannot be used. Therefore, we propose using the integrated complete log-likelihood \citep[ICL;][]{biernacki00}, which relies on the complete data log-likelihood instead of the observed log-likelihood. This criterion is called the ICL--BIC, similar to that used by \cite{jacques17} and is given by
$$
\text{ICL--BIC}=p(\bfx,\hat{\bfz},\hat{\bfwmu},\hat{\bfwsig};\hat{\bvtheta})-\frac{G-1}{2}\log n-\frac{\Lmu+\Lsig-2}{2}\log p-\frac{G(\Lmu+\Lsig)}{2}\log np.
$$
From the property proven by \cite{brault17}, the BIC and ICL--BIC exhibit the same behaviour for large values of $n$ and/or $p$, thus the number of blocks chosen by this criterion is consistent (under some conditions not mentioned here). The model with the largest ICL--BIC is retained.
\paragraph{Search Algorithm}
Because an extra layer of complexity is introduced with the parameter-wise model by considering two column partitions, it may take a very long time to perform an exhaustive search of all possible combinations of $G,\Lmu$ and $\Lsig$ in a pre-defined range. This has been discussed in the literature, specifically by  \cite{robert17}, and a non-exhaustive search algorithm for the parameter-wise model is now presented. Specifically, the algorithm begins with the parameters $(G,\Lmu,\Lsig)=(G_1,\Lmu_1,\Lsig_1)$. Three models with parameters $(G_1+1,\Lmu,\Lsig)$, $(G_1,\Lmu+1,\Lsig)$ and $(G_1,\Lmu,\Lsig+1)$ are then fit. The set with the highest ICL--BIC is retained and we obtain the set $(G_2,\Lmu_2,\Lsig_2)$. The procedure is then repeated until a maximum threshold is reached for these parameters or the ICL--BIC no longer increases. Although not as pertinent for traditional co-clustering, a similar non-exhaustive search algorithm can be used for traditional co-clustering.

\section{Numerical Experiments on Artificial Data}\label{sec:exp}
\subsection{Algorithm and Parameter Estimation Evaluation}
Two different simulations are performed to evaluate the algorithm, parameter estimation, and classification performance.
\subsubsection*{Simulation 1}
50 datasets are simulated according to the following parameters. $n=1000$, $p=100$, $G=3$, $\Lmu=2$, $\Lsig=3$,
$$
{\boldsymbol \mu}=\left(
\begin{array}{ll}
1 &-1\\
2 & -2\\
3 & -3\\
\end{array}
\right), \qquad
{\boldsymbol \Sigma}=\left(
\begin{array}{ccc}
1&0.5 &0.75\\
2 & 1.75 & 0.25\\
1.5 & 2.25 & 2.5\\
\end{array}
\right),
$$
and mixing proportions
$$
\bpi =(0.3,0.3,0.4), \qquad \brhomu=(0.4,0.6), \qquad \brhosig=(0.3,0.3,0.4).
$$
To clarify notation, the cell $g\lmu$ in the matrix $\mu$ corresponds to the mean of an observation from row-cluster $g$ and column-cluster by means $\lmu$, i.e., $\mu_{g\lmu}$. Likewise, the cell $g\lsig$ in the matrix $\matsig$ corresponds to the variance of an observation from row-cluster $g$ and column-cluster by variances $\lsig$, i.e., $\sigma^2_{g\lsig}$.

A burn-in of 20 iterations for the SEM-Gibbs algorithm is used, followed by 100 iterations, followed by 20 iterations of the SE-step to obtain the final partitions.

The error in the mean estimates is calculated using $$\Delta {\boldsymbol\mu}=\sum_{g,\lmu}|\hat{\mu}_{g\lmu}-\mu_{g\lmu}|.$$ The errors for the other parameters are calculated in a similar fashion and are denoted by $\Delta {\boldsymbol \Sigma}$, $\Delta {\boldsymbol \pi}$, $\Delta {\boldsymbol \rho}^{\mu}$ and $\Delta {\boldsymbol \rho}^{\Sigma}$, respectively. The averaged errors (and their standard deviations) over the 50 datasets are shown in Table \ref{tab:sim1aest}.  The average errors are low for all variables indicating good parameter recovery.

The adjusted rand index \citep[ARI;][]{hubert85} is used to assess classification performance. This quantity compares two partitions, in this case the true partition to an estimated partition, and has a value of 1 if there is perfect agreement, and an expected value of 0 under random classification. Table~\ref{tab:sim1aARI} displays the average ARI, with standard deviations, for the row, column by means, and column by variances partitions over the 50 simulated datasets. Notice that the classification is perfect for both partitions by columns for all simulated datasets. Moreover, the average ARI for the rows is very high.
\begin{table}[!htb]
\centering
\caption{Average error (and standard deviation) of the parameter estimates over the 50 datasets for Simulation 1.}
\begin{tabular}{ccccc}
\hline
$\overline{\Delta {\boldsymbol \mu}}$ & $\overline{\Delta {\boldsymbol \Sigma}}$ & $\overline{\Delta  {\boldsymbol\pi}}$ &$\overline{\Delta \brhomu}$&$\overline{\Delta \brhosig}$\\
\hline
0.14 (0.70)&0.24 (0.75) & 0.012 (0.082) & 1.44e-15 (5.61e-16)&1.33e-15 (4.59e-16)\\
\hline
\end{tabular}
\label{tab:sim1aest}
\end{table}
\begin{table}[!htb]
\centering
\caption{Average ARI (and standard deviation) for the row ($\overline{\text{ARI}}_r$), column by means ($\overline{\text{ARI}}_{c \mu}$), and column by variances ($\overline{\text{ARI}}_{c \Sigma}$) partitions over the 50 datasets for Simulation 1.}
\begin{tabular}{ccc}
\hline
$\overline{\text{ARI}}_r$ & $\overline{\text{ARI}}_{c \mu}$ & $\overline{\text{ARI}}_{c \Sigma}$\\
\hline
0.99 (0.068)&1.00 (0.00)&1.00 (0.00)\\
\hline
\end{tabular}
\label{tab:sim1aARI}
\end{table}

\begin{figure}[!htb]
\centering
\includegraphics[width=0.9\textwidth]{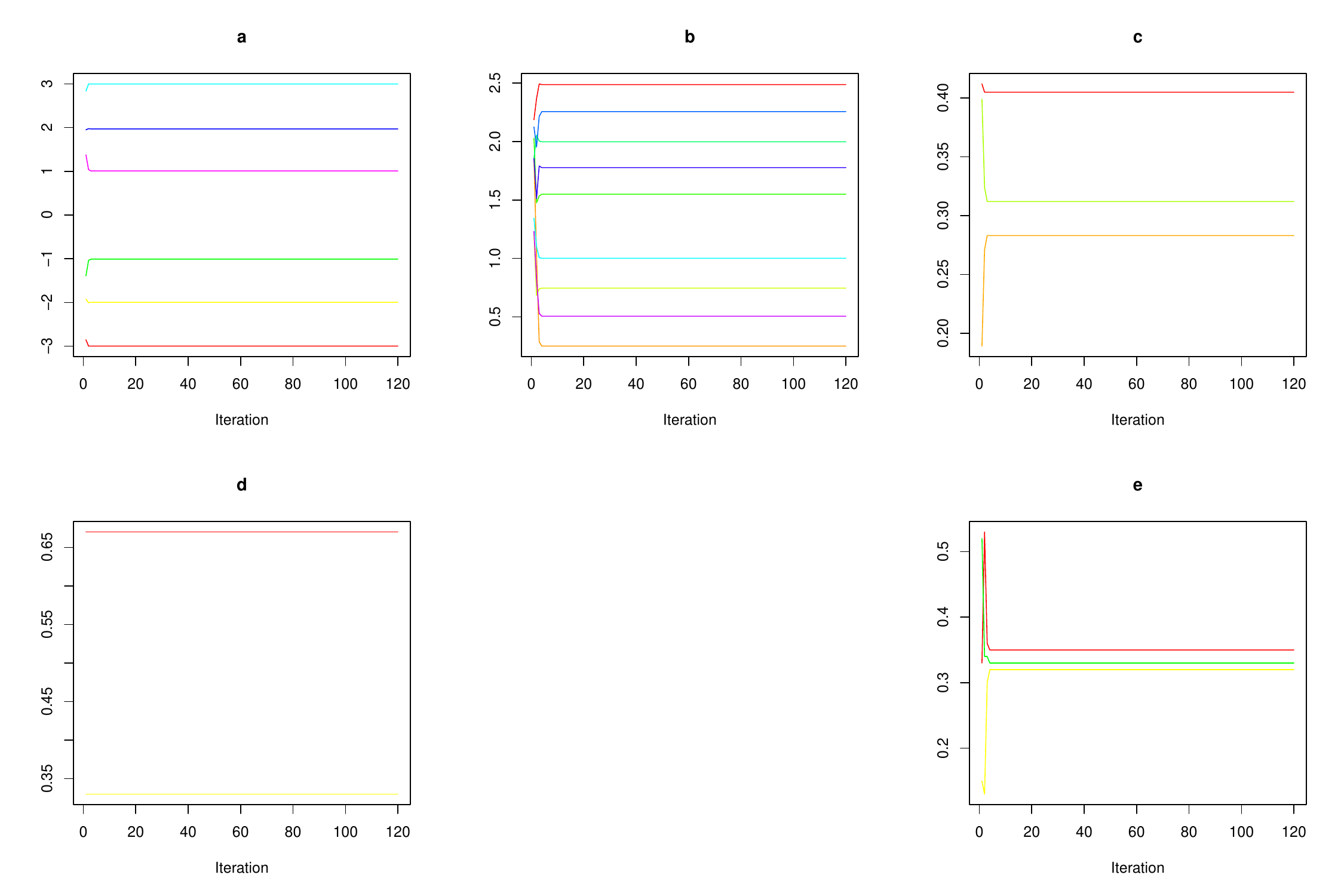}
\caption{SEM algorithm parameter estimation progression for one dataset for (a) the mean parameters $\mu_{g\lmu}$, (b) the variance parameters $\sigma^2_{g\lsig}$, (c) the row mixing proportions $\pi_g$, (d) the column by means mixing proportions $\rhomu_{\lmu}$, and (e) the column by variances mixing proportions $\rhosig_{\lsig}$ for Simulation~1.}
\label{fig:Sim1est}
\end{figure}

In Figure \ref{fig:Sim1est}, the progression of the parameter estimates over the course of the SEM-Gibbs algorithm is shown for one of the datasets (the other datasets exhibit similar behaviour).  From these plots, it is clear that a burn-in of 20 iterations is sufficient to obtain a stable chain.

\begin{figure}[!htb]
\centering
\includegraphics[width=0.75\textwidth]{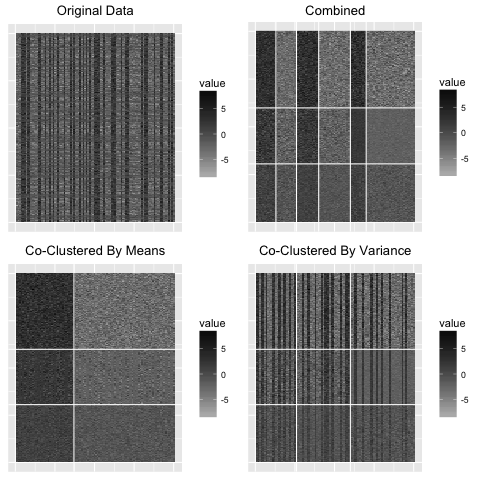}
\caption{Estimated co-clustering solution for one of the fifty datasets from Simulation 1.}
\label{fig:coclustsim1}
\end{figure}

Finally, in Figure \ref{fig:coclustsim1}, the co-clustering results for one of the 50 datasets is displayed. Note, in this case, the estimated co-clustering result is the same as the true co-clustering solution. In the top left panel, a heatmap of the original data is displayed. In the co-clustering by means panel (bottom left), the co-clustering results for the row-clusters and the column-clusters by means is shown. The co-clustering by variances panel (bottom right) shows the co-clustering results for the row-clusters and the column-clusters by variances. Finally, the combined co-clustering (top right) displays the co-clustering solution with all combined column-clusters. Specifically, going from left to right, the first combined column-cluster consists of the columns partitioned into column-cluster 1 for the means and column-cluster 1 for the variances, the second combined column-cluster are the columns clustered into column-cluster 2 for the means and column-cluster 1 for the variances and so on. Combining the column-clusters by means and variances in this manner results in a maximum of $\Lmu\Lsig$ combined column-clusters (as is the case here) thus allowing more flexibility. It is important to note, however, that there may be cases, as we will see with the real dataset, when no columns are clustered into a particular pair $\lmu$ and $\lsig$, and thus the combined co-clustering result might have fewer than $\Lmu\Lsig$ combined column-clusters but never more.

\subsubsection*{Simulation 2}
\begin{figure}[!htb]
\centering
\includegraphics[width=0.9\textwidth]{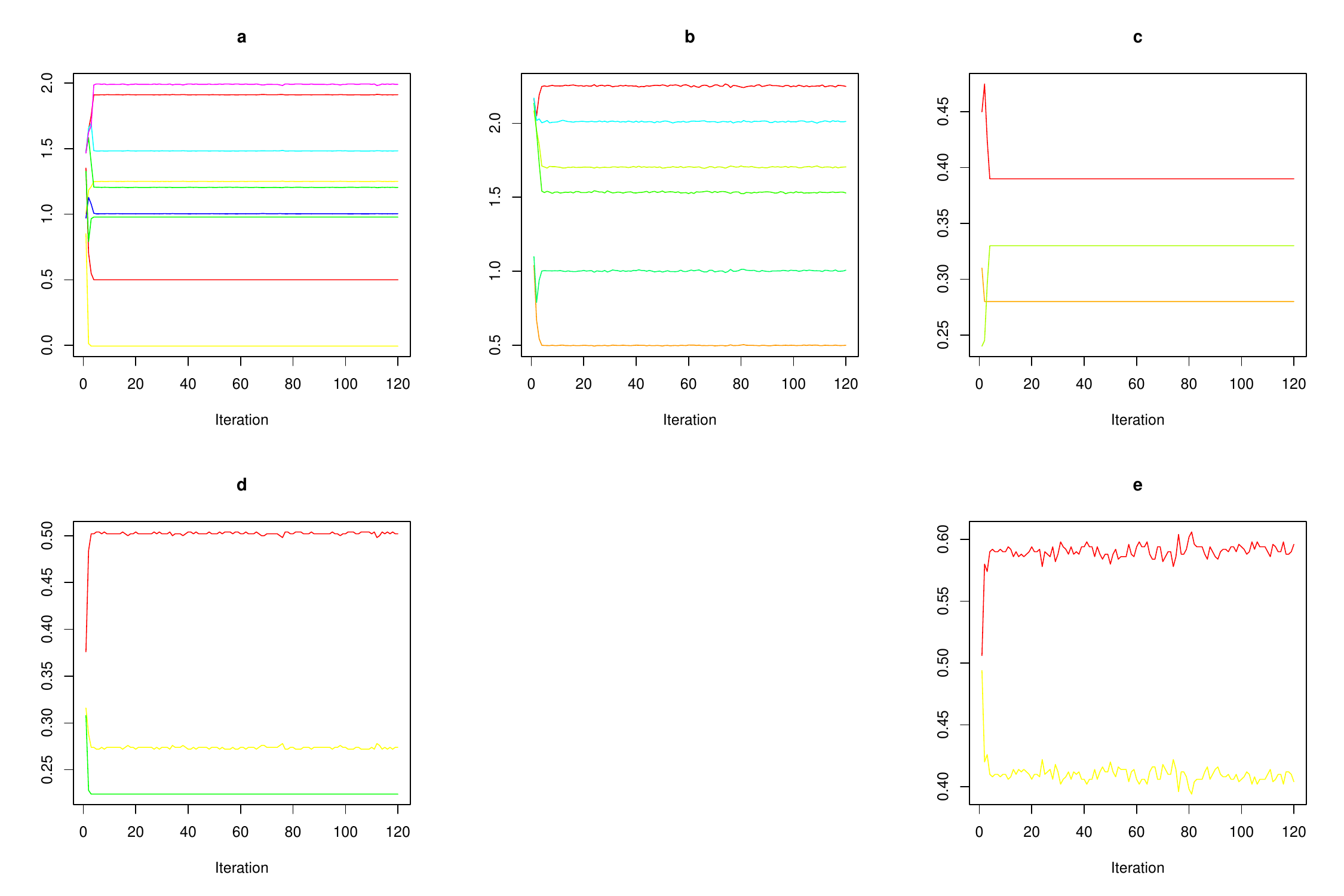}
\caption{Simulation 2 SEM algorithm parameter estimation progression for one dataset for (a) the mean parameters $\mu_{g\lmu}$, (b) the variance parameters $\sigma^2_{g\lsig}$, (c) the row mixing proportions $\pi_g$, (d) the column by means mixing proportions $\rhomu_{\lmu}$, and (e) the column by variances mixing proportions $\rhosig_{\lsig}$.}
\label{fig:Sim2est}
\end{figure}

\begin{figure}[!htb]
\centering
\includegraphics[width=0.65\textwidth]{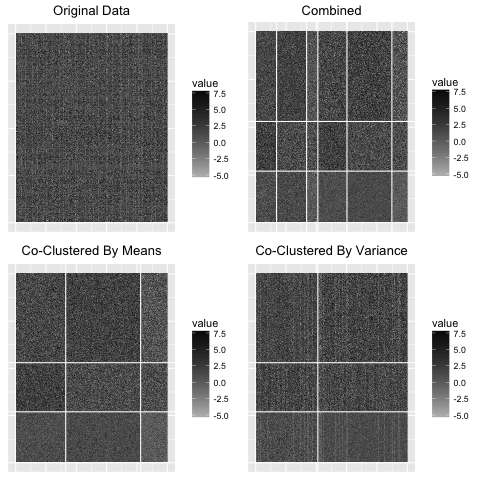}
\caption{Estimated co-clustering solution for one of the fifty datasets from Simulation 2.}
\label{fig:coclustsim2}
\end{figure}

In Simulation 2, less separation between groups is considered. A total of 50 datasets are again considered with the parameters $n=200$, $p=500$, $G=3$, $\Lmu=3$, $\Lsig=2$,
$$
{\boldsymbol \mu}=\left(
\begin{array}{lll}
1 &1.25&0\\
2 & 1.2&1\\
1.5 & 1.9&0.5\\
\end{array}
\right), \qquad 
{\boldsymbol \Sigma}=\left(
\begin{array}{cc}
1&0.5\\
2 & 1.75\\
1.5 & 2.25\\
\end{array}
\right),
$$
and the mixing proportions
$$
{\boldsymbol \pi}=(0.3,0.3,0.4), \qquad \brhomu=(0.3,0.5,0.2), \qquad \brhosig=(0.4,0.6).
$$
Table \ref{tab:sim1best} shows the average error of the estimates over the 50 datasets, and the average ARI values over the 50 datasets for each partition are shown in Table  \ref{tab:sim1bari}. Again, we obtain very good classification performance for all three partitions. The progression of the parameter estimates is shown in Figure \ref{fig:Sim2est}. Similar to Simulation 1, a burn-in period of 20 iterations is still sufficient to obtain a stable chain.  Finally, Figure \ref{fig:coclustsim2} displays the co-clustering solutions for one of the 50 datasets. Unlike in the first simulation, there is very little spatial separation between blocks. 
\begin{table}[!htb]
\centering
\caption{Average error (and standard deviation) of the estimates over the 50 datasets for Simulation 2.}
\begin{tabular}{ccccc}
\hline
$\overline{\Delta {\boldsymbol \mu}}$ & $\overline{\Delta {\boldsymbol \Sigma}}$ & $\overline{\Delta  {\boldsymbol\pi}}$ &$\overline{\Delta \brhomu}$&$\overline{\Delta \brhosig}$\\
\hline
0.15 (0.50)&0.085 (0.046) & 1.29e-15 (3.91e-16) & 0.015 (0.088)&0.0079 (0.0054)\\
\hline
\end{tabular}
\label{tab:sim1best}
\end{table}
\begin{table}[!htb]
\centering
\caption{Average ARI (and standard deviation) for the row ($\overline{\text{ARI}}_r$), column by means ($\overline{\text{ARI}}_{c \mu}$), and column by variances ($\overline{\text{ARI}}_{c \Sigma}$) partitions over the 50 datasets for Simulation 2.}
\begin{tabular}{ccc}
\hline
$\overline{\text{ARI}}_r$ & $\overline{\text{ARI}}_{c \mu}$ & $\overline{\text{ARI}}_{c \Sigma}$\\
\hline
1.00 (0.00)&0.98 (0.080)&0.96 (0.018)\\
\hline
\end{tabular}
\label{tab:sim1bari}
\end{table}

\subsection{Simulation 3}
In this simulation, the performance of the ICL--BIC selection criterion is considered. Again, 50 datasets are simulated with  $n=2000$, $p=500$, $G=\Lmu=\Lsig=3$,
\\
$$
{\boldsymbol \mu}=\left(
\begin{array}{lll}
1 &1.25&0\\
2 & 1.2&1\\
1.5 & 1.9&0.5\\
\end{array}
\right), \qquad 
{\boldsymbol \Sigma}=\left(
\begin{array}{ccc}
1&0.5&0.25\\
2 & 1.75&0.5\\
1.5 & 2.25&1\\
\end{array}
\right),
$$
and mixing proportions
$$
\bpi=(0.3,0.3,0.4), \qquad \brhomu=(0.3,0.4,0.3), \qquad \brhosig=(0.4,0.3,0.3).
$$
An exhaustive search is performed considering each of combination of $G,\Lmu,\Lsig\in\{2,3,4\}$. In Table \ref{tab:SimICL}, the number of times each value of $G$, $\Lmu$ and $\Lsig$ is chosen by the ICL--BIC is displayed. For the vast majority of the datasets, the correct model is chosen by the ICL--BIC.
\begin{table}[!htb]
\centering
\caption{Frequency of the number of row-clusters, column-clusters by means, and column-clusters by variances chosen by the ICL--BIC over the 50 simulated datasets when using the exhaustive search in Simulation 3.}
\begin{tabular}{cccc}
\hline
& 2&3&4\\
\hline
$G$ &0&49&1\\
$\Lmu$& 0 & 48 & 2\\
$\Lsig$&0 & 48 & 2\\
\hline
\end{tabular}
\label{tab:SimICL}
\end{table}

\subsection{Simulation 4}
In the last simulation, the performance of the non-exhaustive search algorithm described in Section 3.3 is addressed. In all, 25 datasets are simulated according to the parameters $n=100, p=200, G=\Lsig=3, \Lmu=4$,
$$
{\boldsymbol \mu}=\left(
\begin{array}{cccc}
1& -0.25 & 0.3& -1\\
1.25& 0& 0.1& -0.3\\
0.5& -1&  0&  0.1\\
\end{array}
\right), \qquad 
{\boldsymbol \Sigma}=\left(
\begin{array}{ccc}
1&0.5&0.25\\
2 & 1.75&0.5\\
1.5 & 2.25&1\\
\end{array}
\right),
$$
and
$$
{\boldsymbol \pi}=(0.3,0.3,0.4), \qquad \brhomu=(0.2,0.3,0.25,0.25), \qquad \brhosig=(0.5,0.25,0.25).
$$
The initial values are taken to be $(G_1,\Lmu_1,\Lsig_1)=(1,1,1)$ and the maximum values for all three are set to five. In Table~\ref{tab:nonexsim}, the number of times each value of $G$, $\Lmu$ and $\Lsig$ is chosen by the ICL--BIC is shown. Notice that the procedure performs quite well for choosing the correct model.

\begin{table}[!htb]
\centering
\caption{Frequency of the number of row-clusters, column-clusters by means, and column-clusters by variances chosen by the ICL--BIC over the 25 simulated datasets when using the non-exhaustive search method for Simulation 4.}
\begin{tabular}{cccc}
\hline
& 2&3&4\\
\hline
$G$ &0&24&1\\
$\Lmu$& 0 & 0 & 25\\
$\Lsig$&1 & 24 & 0\\
\hline
\end{tabular}
\label{tab:nonexsim}
\end{table}

\section{Real Data Analyses}
\subsection{Comparing Parameter-Wise and Traditional Co-Clustering Under Similar Conditions}
A subset of the Jester dataset used by \cite{goldberg01} is used to compare parameter-wise co-clustering and traditional co-clustering. The data consist of 100 jokes rated on a ``continuous" scale from $-10$ to $10$. A total of 7200 users rated all 100 jokes, and a random sample of 2000 of these users is considered herein.

\begin{figure}[!htb]
\centering
\includegraphics[width=0.65\textwidth]{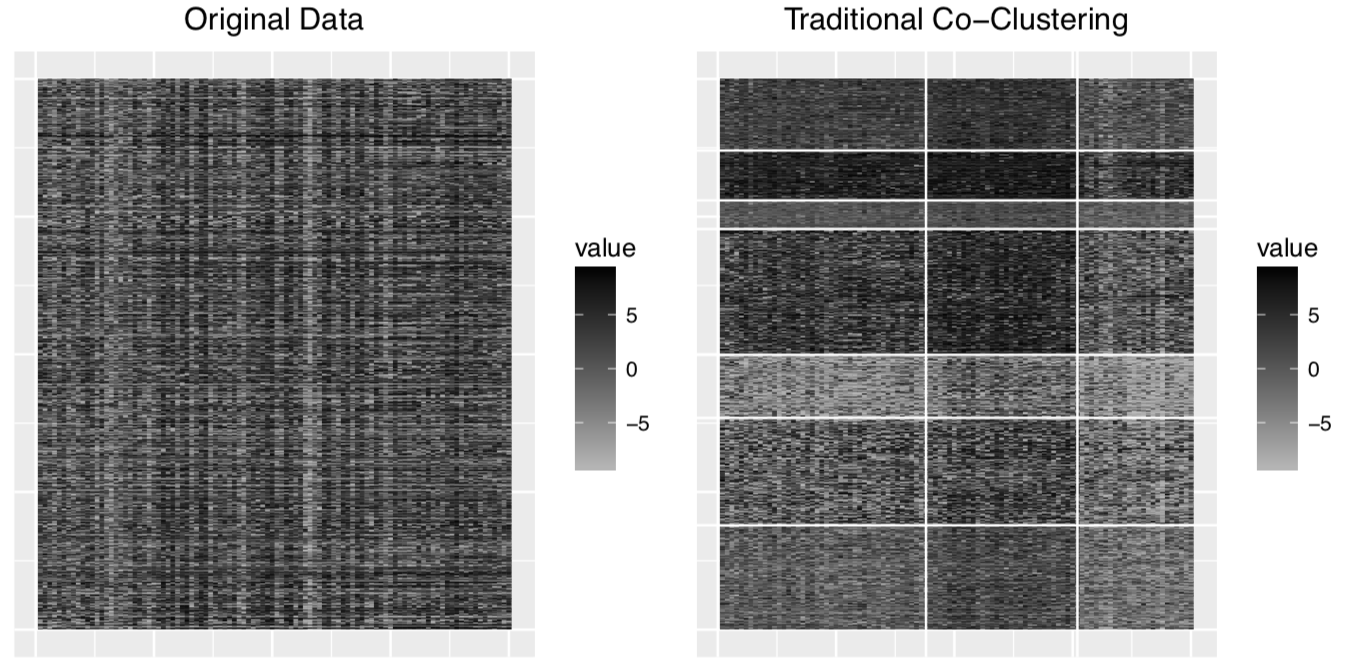}
\caption{Traditional co-clustering results for the Jester data.}
\label{fig:jokes_nonext}
\end{figure}

\begin{figure}[!htb]
\centering
\includegraphics[width=0.65\textwidth]{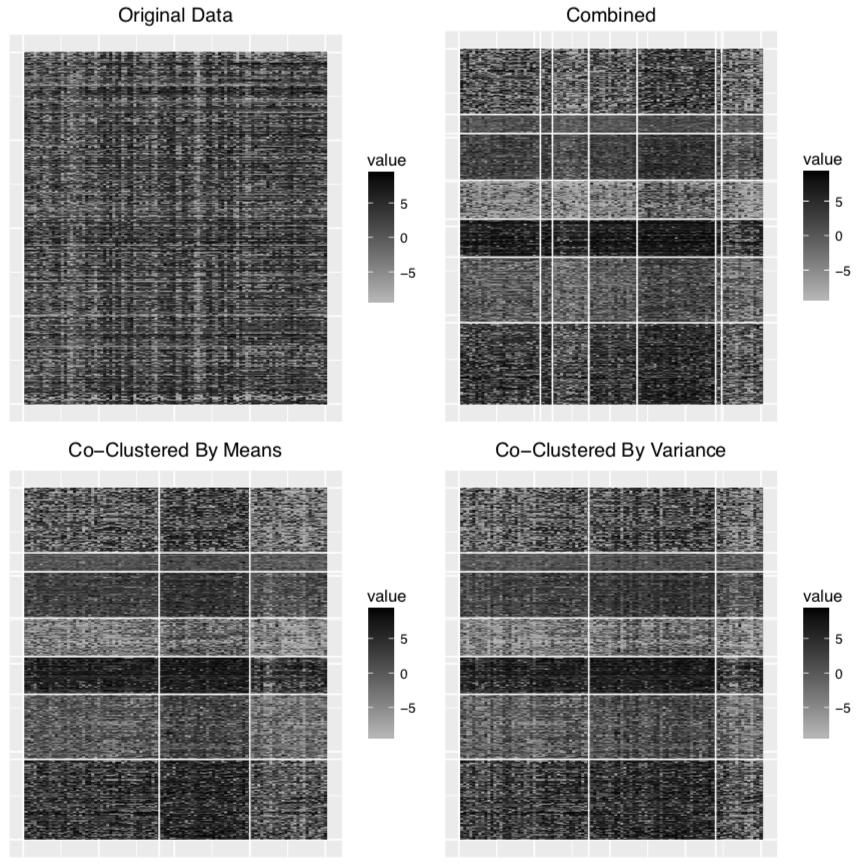}
\caption{Parameter-wise co-clustering results for the Jester dataset under similar conditions to the traditional co-clustering solution.}
\label{fig:jokes_nonexs}
\end{figure}

The non-exhaustive search algorithm is performed for traditional co-clustering with the number of row-clusters ranging from one to 25 and the number of column-clusters ranging from one to seven. This results in choosing seven row-clusters and three column-clusters and the resultant ICL--BIC is $-569487.0$. With these values for $G$ and $L$, the total number of free parameters is 50. In the next section, the non-exhaustive search algorithm is used for the proposed parameter-wise method; however, it is interesting to consider the performance of the parameter-wise method under similar conditions to the results obtained with traditional co-clustering. Specifically, the parameter-wise method is performed on this dataset with $G=7, \Lmu=\Lsig=3$. Under this model, the ICL--BIC is $-569010.4$, and the total number of free parameters is 52. Note that the ICL--BIC values for both traditional and parameter-wise co-clustering are quite similar, with a slightly higher value obtained when using parameter-wise co-clustering. In Figure \ref{fig:jokes_nonext}, the original data (left panel) and the traditional co-clustering solution (right panel), are shown, and the co-clustering solutions for parameter-wise co-clustering are displayed (Figure~\ref{fig:jokes_nonexs}) in the same format as the simulations. Notice that a total of seven combined column-clusters are obtained when using parameter-wise co-clustering. 

In Table \ref{tab:nvtmeans}, we show a classification table comparing the column-clusters by means and column-clusters by variances found using parameter-wise co-clustering and the column-clusters found using traditional co-clustering. There is almost perfect agreement between the column-clusters from traditional co-clustering and the column-clusters by means from parameter-wise co-clustering. This, however, is not true for the column-clusters by variances. This result is somewhat perceptible in the images of the co-clustering solutions. In Table~\ref{tab:nvtG}, the classification table comparing row-clusters from traditional and parameter-wise co-clustering is displayed. It is clear that the row-clusters found by both of these methods are quite comparable --- the ARI when comparing these two partitions is 0.86.
\begin{table}[ht]
\centering
\caption{Classification table comparing the column-clusters by means and column-clusters by variances for parameter-wise co-clustering and column-clusters from traditional co-clustering for the Jester dataset.}
\begin{tabular}{cccc|ccc}
&\multicolumn{3}{c}{Means}&\multicolumn{3}{c}{Variances}\\
  \hline
Traditional & 1 & 2 & 3& 1&2&3 \\ 
  \hline
1 &  43 &   0 &   1& 28 &  14 &   2 \\
  2 &   2 &  30 &   0&   4 &  28 &   0 \\ 
  3 &   0 &   0 &  24  &  11 &   0 &  13 \\ 
   \hline
\end{tabular}
\label{tab:nvtmeans}
\end{table}
\begin{table}[ht]
\centering
\caption{Classification table comparing row-clusters for parameter-wise and traditional co-clustering.}
\begin{tabular}{rrrrrrrr}
& \multicolumn{7}{c}{Traditional}\\
  \hline
Parameter-Wise & 1 & 2 & 3 & 4 & 5 & 6 & 7 \\ 
  \hline
1 & 427 &  10 &   1 &   0 &   3 &   0 &  16 \\ 
  2 &   0 & 350 &   0 &   9 &   0 &   0 &  11 \\ 
  3 &  18 &   0 & 180 &   0 &  16 &   0 &   0 \\ 
  4 &   0 &   0 &   0 & 216 &   0 &   0 &   3 \\ 
  5 &  10 &  11 &   0 &   0 & 241 &   1 &   0 \\ 
  6 &   0 &   5 &   0 &   0 &   0 & 103 &   0 \\ 
  7 &   2 &   3 &   0 &   4 &   0 &   0 & 360 \\ 
   \hline
\end{tabular}
\label{tab:nvtG}
\end{table}

\subsection{Further Analysis with Parameter-Wise Co-Clustering}
\begin{figure}[!htb]
\centering
\includegraphics[width=0.65\textwidth]{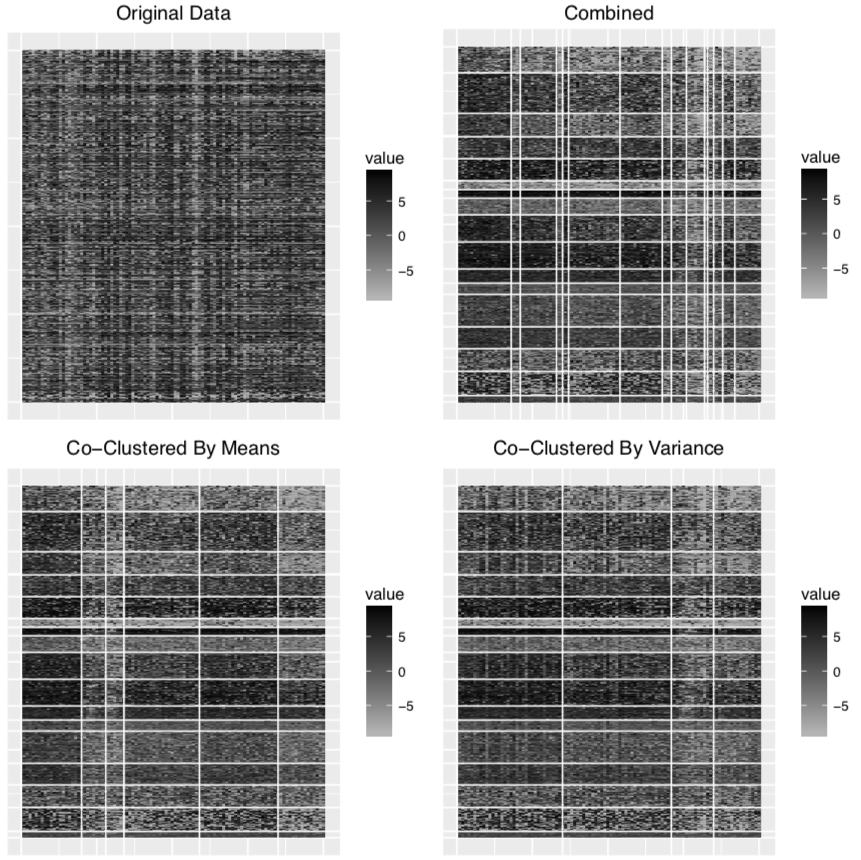}
\caption{Parameter-wise co-clustering results for the Jester data after performing the non-exhaustive search algorithm.}
\label{fig:jokes_nonexne}
\end{figure}

The non-exhaustive search algorithm is now performed for parameter-wise co-clustering. The range of values was one to 25 row-clusters, and one to seven column-clusters by means and column-clusters by variances resulting in the ICL--BIC choosing a model with 17 row-clusters, six column-clusters by means, and four column-clusters by variances.  The resulting ICL--BIC is $-561099.0$ and a total of 15 combined column-clusters are obtained. Notice that there is significant improvement in the ICL--BIC in this case. In Figure~\ref{fig:jokes_nonexne}, we show the parameter-wise co-clustering solution. Because more row-clusters are obtained, it is far more difficult to visualize the row-clusters. Moreover, the combined co-clustering solution is very difficult to interpret in this scenario, which displays the benefit of visualizing the column-clusters by means and column-clusters by variances separately.
 \begin{figure}[!htb]
 \centering
 \includegraphics[height=0.55\textwidth]{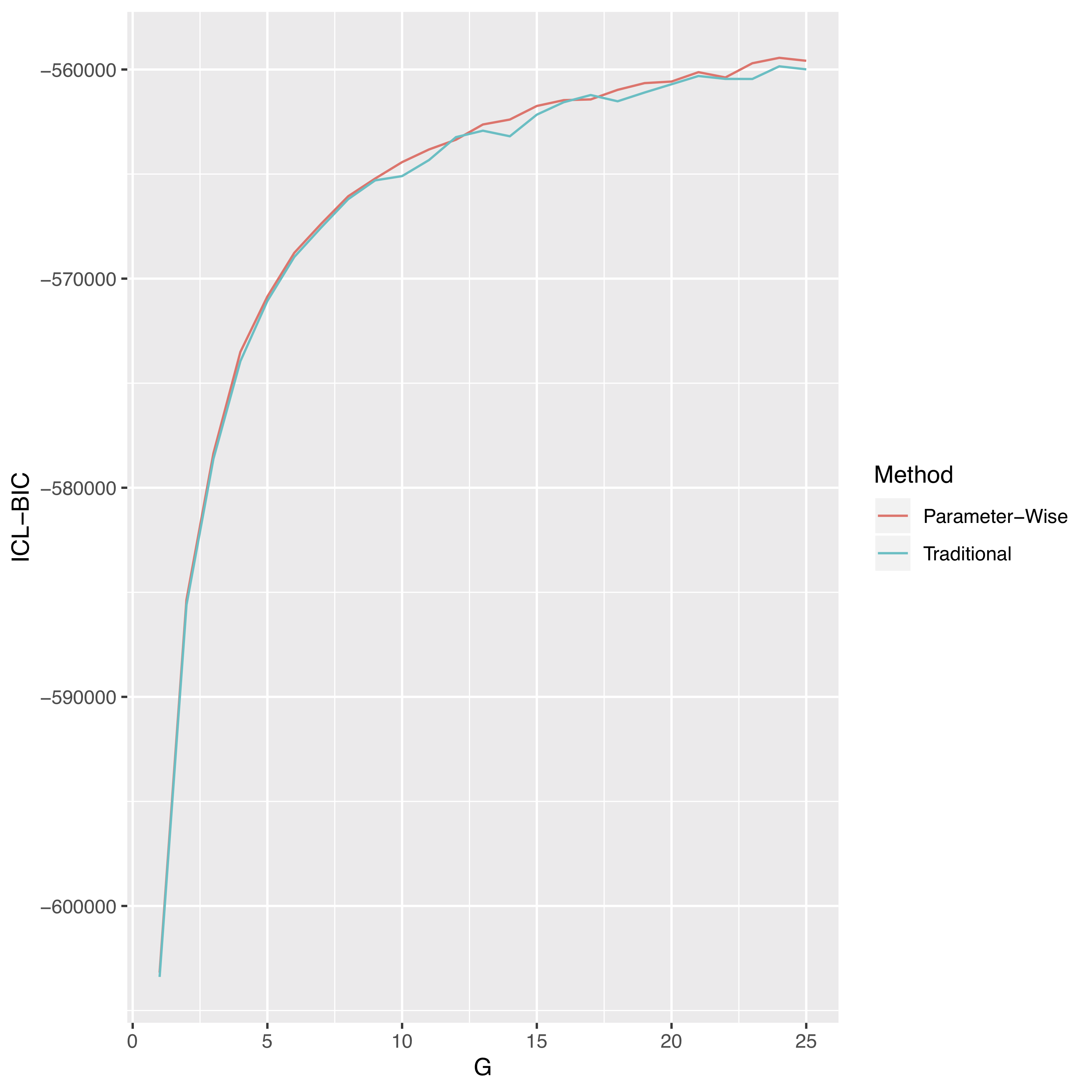}
  \caption{Maximum ICL--BIC over $L$ for traditional co-clustering (turquoise), and $\Lmu$ and $\Lsig$ for parameter-wise co-clustering (red) for each value of $G$, against $G$.}
 \label{fig:ICLres}
 \end{figure}

Finally, the exhaustive search algorithm is performed for both traditional and parameter-wise co-clustering. For each value of $G\in\{1,2,\ldots,25\}$, the maximum ICL--BIC over all values of $L$ for traditional co-clustering, and $\Lmu$ and $\Lsig$ for parameter-wise co-clustering is considered. In Figure \ref{fig:ICLres}, we display a plot of this maximum ICL--BIC against $G$. For both traditional and parameter-wise co-clustering, the ICL--BIC begins to plateau around $G=10$. Moreover, the ICL--BIC for parameter-wise co-clustering is oftentimes, if only very slightly, higher than traditional co-clustering. Finally, we note that it is very computationally expensive to run the exhaustive search with parameter-wise co-clustering taking around 24 hours using 25 1200MHz cores running continuously.

\section{Discussion}
A parameter-wise co-clustering algorithm was developed for high-dimensional data. This parameter-wise method allowed for two partitions of the columns based on both means and variances, as well as a combined co-clustering solution. This, in essence, provides more flexibility than traditional co-clustering, while maintaining the high degree of parsimony inherent to traditional co-clustering. An SEM Gibbs algorithm was used for parameter estimation, and evaluated by two simulations. An ICL--BIC criterion, as well as a non-exhaustive search algorithm, were developed for model selection.

A subset of the Jester dataset was considered for comparison purposes between traditional and parameter-wise co-clustering. After applying traditional co-clustering to the data, parameter-wise co-clustering was performed using similar parameters, i.e., same $G$ and $\Lmu=\Lsig=L$. This resulted in similar row-clusters between the two methods. Furthermore, the column-clusters by means using parameter-wise co-clustering were almost identical to the column-clusters from traditional co-clustering.  This was not true, however, when comparing the column-clusters by variances and the column-clusters obtained from traditional co-clustering. Parameter-wise co-clustering also had a marginally higher ICL--BIC in this case. Using the non-exhaustive search algorithm for parameter-wise co-clustering resulted in far more row-clusters, and many more combined column-clusters, which displayed the utility of considering the co-clustering by means, and co-clustering by variances separately from the combined co-clustering solution.

Although this method only considered the use of the Gaussian distribution, it can be extended in various ways. One example would be to use other continuous distributions with more than one parameter. For example, one could consider the skew-$t$ distribution and cluster columns based on location, scale, concentration and skewness. This could also be extended to data that cannot be considered a realization of a continuous random variable such as ordinal data where the columns could be partitioned according to mode and precision. The number of free parameters in each of these cases will not depend on the dimensionality of the data thus preserving the parsimony inherent to co-clustering. 


\end{document}